\documentclass{article} 
\usepackage{colm2024_conference}

\usepackage{microtype}
\usepackage{hyperref}
\usepackage{url}
\usepackage{booktabs}
\definecolor{darkblue}{rgb}{0, 0, 0.5}
\hypersetup{colorlinks=true, citecolor=darkblue, linkcolor=darkblue, urlcolor=darkblue}
\usepackage{booktabs} 
\usepackage{tabularx}   
\usepackage{enumitem}
\usepackage{graphicx}
\usepackage{url}
\usepackage{bigfoot}
\usepackage{booktabs}
\usepackage{multirow}
\usepackage[most]{tcolorbox}
\usepackage{graphicx}
\usepackage{subcaption}
\usepackage{etoolbox}
\usepackage{booktabs}
\usepackage{bbding} 
\let\classAND\AND
\let\AND\relax
\usepackage{algorithmic}

\let\AND\classAND
\AtBeginEnvironment{algorithmic}{\let\AND\algoAND}

\usepackage{algorithm}
\usepackage{algorithmic}

\usepackage{soul}
\usepackage{url}
\usepackage{booktabs}
\usepackage{arydshln}
\usepackage{nicefrac}
\usepackage{multirow}
\usepackage{tabularx}
\usepackage{amsmath}
\usepackage{amssymb}
\usepackage{amsfonts}
\usepackage{mathtools}
\usepackage{fontawesome}

\usepackage{xcolor,colortbl}
\usepackage[export]{adjustbox}

\usepackage{wrapfig}

\tcbuselibrary{listings,breakable}
\tcbset{listing engine=listings,colframe=black,colback=white,size=small}

\usepackage{upquote}
\definecolor{dkgreen}{rgb}{0,0.6,0}
\definecolor{gray}{rgb}{0.5,0.5,0.5}
\definecolor{mauve}{rgb}{0.58,0,0.82}
\usepackage{xspace}

\usepackage{setspace}

\tcbset{
  promptstyle/.style={
    colback=gray!15,   
    colframe=gray!60,  
    boxrule=0.4pt,     
    arc=2pt,           
    left=4pt,right=4pt,top=4pt,bottom=4pt,  
    fonttitle=\bfseries,
  }
}
\newtcolorbox{promptblock}[1][]{promptstyle,#1}

\usepackage{fvextra}
\RecustomVerbatimEnvironment{verbatim}{Verbatim}{breaklines=true, breakanywhere=true}

\newcommand{\RealMirror}{\raisebox{-4.5pt}{\includegraphics[height=1.5em]{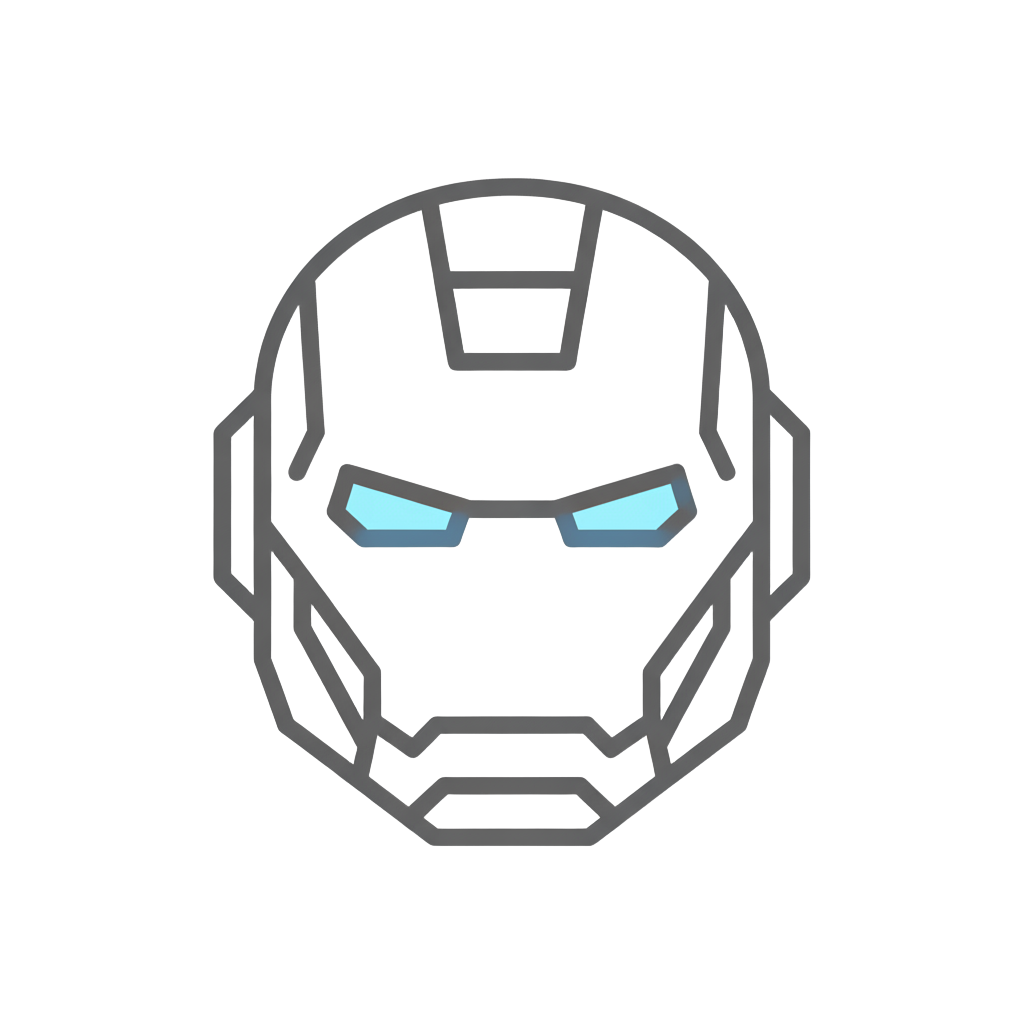}}\hspace{-7.0pt}\xspace}

\newcommand\blfootnote[1]{
    \begingroup
    \renewcommand\thefootnote{}\footnote{#1}
    \addtocounter{footnote}{-1}
    \endgroup
}

\title{\LARGE MirrorLimb: Implementing hand pose acquisition and robot teleoperation based on RealMirror}

\author{
\vspace{-4mm}
\\
\normalsize{}
Cong Tai\dag\hspace{3mm} 
Hansheng Wu\dag\hspace{3mm}
Haixu Long\hspace{3mm}
Zhengbin Long\hspace{3mm}
\\
\normalsize{}
Zhaoyu Zheng\hspace{3mm}
Haodong Xiang\hspace{3mm}
Tao Shen*\hspace{3mm}
\\
\vspace{20mm}
\textbf{ZTE Terminators Group}
\vspace{-15mm}
}

\colmfinalcopy
\begin{document}

\maketitle

\begin{center}
\vspace{-1cm}
\begin{tabular}{rl}
\RealMirror & \url{https://github.com/terminators2025/RealMirror} \\
\end{tabular}
\end{center}

\begin{abstract}
In this work, we present a PICO-based robot remote operating framework that enables low-cost, real-time acquisition of hand motion and pose data, outperforming mainstream visual tracking and motion capture solutions in terms of cost-effectiveness. The framework is natively compatible with the RealMirror ecosystem, offering ready-to-use functionality for stable and precise robotic trajectory recording within the Isaac simulation environment, thereby facilitating the construction of Vision-Language-Action (VLA) datasets. Additionally, the system supports real-time teleoperation of a variety of end-effector-equipped robots, including dexterous hands and robotic grippers. This work aims to lower the technical barriers in the study of upper-limb robotic manipulation, thereby accelerating advancements in VLA-related research.
\end{abstract}

\blfootnote{$^{\dag}$ Equal contribution}
\blfootnote{$^{*}$ Corresponding author. Emails: shen.tao5@zte.com.cn}

\section{Introduction}

\label{sec:intro}
Recently, Vision-Language-Action (VLA) models have become a research hotspot in the field of robotics. Many of these models demonstrate impressive upper-limb manipulation capabilities in robots \citep{zhao2023learning,chi2023diffusion,figure_helix,black2024pi_0,shukor2025smolvla}. To enable more systematic evaluation and evolution of VLA models, related work—RealMirror provides an end-to-end VLA research platform integrating data collection, training, and evaluation \citep{tai2025realmirror}. Within RealMirror, teleoperation data from multiple simulation scenarios serves as the cornerstone of the workflow. Experimental results indicate that the quality and diversity of human demonstrations directly determine the performance of VLA models \citep{yang2025beyond}.

Currently, several mainstream paradigms exist for mapping human motions to robot embodiments. One approach involves using professional motion capture systems: motion data captured from sensors is processed in real time and directly mapped to the corresponding joints of the robot. This method can capture joint angle ground truths with high fidelity, but high hardware costs and issues such as joint drift limit its widespread adoption. Another approach relies on the interactive capabilities of XR or visual devices \citep{cheng2024open,qin2023anyteleop}, mapping motion data from handle or human hands. For XR teleoperation schemes using handles, while the end-pose of the handle can be acquired stably, the grasping actions of the robot are often limited to predefined functional packages, resulting in relatively monotonous movements and a lack of fine manipulation capabilities. For vision-based XR mapping schemes, such as the Apple Vision Pro teleoperation system integrated into the IsaacLab ecosystem \citep{IsaacLabCloudXRTeleoperation}, although expensive wearable hardware is avoided, the Apple Vision Pro remains costly. Moreover, acquiring stable and accurate real-time hand joint poses remains a significant challenge.
To address these challenges, we have extended the capabilities of RealMirror’s handle-based teleoperation by proposing MirrorLimb—a control system based on PICO devices for robot teleoperation via hand gestures or handles. The main contributions of this system are as follows:

1)Native compatibility with the RealMirror platform. Combined with RealMirror’s kinematic/dynamic optimization capabilities for robot motion resolution, it enables highly precise and stable remote robot teleoperation within the IsaacSim simulation environment.

2)Integration of a WebXR/OpenXR-based communication framework and an end-to-end teleoperation software system—providing standardized interfaces for mapped raw data from handle/gesture inputs, facilitating adaptation to different operating platforms and various robot end-effectors.

3)MirrorLimb supports XR devices such as PICO 3 \& 4, offering significantly more cost-effective hardware compared to motion capture systems or Apple Vision Pro, thereby substantially lowering the barrier to achieving fine-grained robotic manipulation.
\section{Data Acquisition And Kinematic Optimization}
\label{sec:model_arc}

\begin{figure}[H]
\centering
\includegraphics[width=\linewidth]{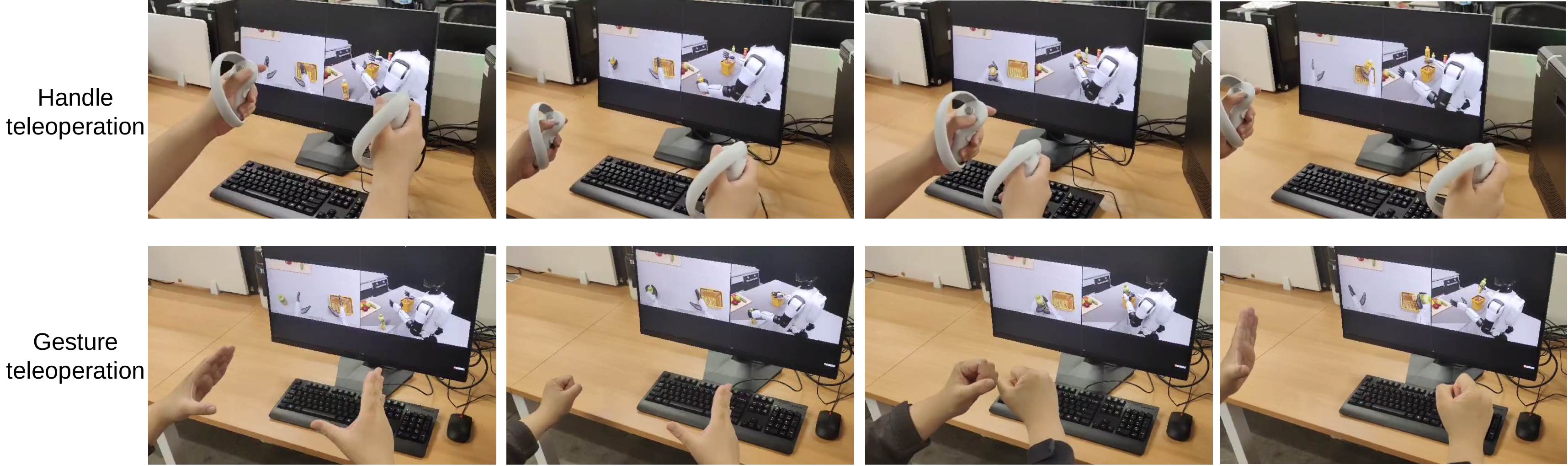}
\caption{Handle and gesture teleoperation demonstration.}
\label{fig:hand&handle}
\end{figure}

We design a dual-channel acquisition stack that captures both handle commands and fine-grained hand gestures from PICO XR devices in real time. The handle path is implemented with WebXR in a browser runtime and streamed to a secure Node.js server, while the gesture path uses OpenXR (via PICO’s Unity SDK) and transmits fixed-rate joint poses via UDP. Both streams are normalized to the RealMirror/IsaacSim coordinate conventions and exposed through standardized schemas for downstream VLA data curation and teleoperation. The handle and gesture teleoperation diagram is shown in Figure \ref{fig:hand&handle}.

\subsection{Architectural Overview}
\begin{itemize}
\item The acquisition pipeline prioritizes low latency, stability, and cost-efficiency.

\item WebXR handle data are sampled in the XR frame loop without rendering overhead, quantized, and pushed over a persistent socket to a server that performs atomic file updates for consumers in IsaacSim/RealMirror.

\item OpenXR hand joint poses are sampled at a fixed 60 Hz independent of rendering, serialized into a compact binary payload, and sent via UDP to minimize end-to-end latency.

\end{itemize}

\subsection{WebXR-Based Handle Acquisition}
The handle acquisition path is implemented within a browser runtime using WebXR, prioritizing efficiency and robustness. To minimize system overhead, an XR session is initiated without performing any scene rendering; its frame loop is used exclusively to drive data capture at the device's native refresh rate (typically 72–90 Hz). For each frame, the system queries the 6-DoF poses for both the grip space (handle body) and the target ray space (pointer), alongside the state of all buttons and joystick axes.

A critical step in this process is coordinate system harmonization. Poses originating from the WebXR runtime (a right-handed, Y-up frame) are transformed into IsaacSim’s right-handed, Z-up world convention. This involves a consistent change of basis for both position and orientation to ensure seamless integration. To enhance control stability and reduce network bandwidth, position data is quantized to 1 mm resolution. The resulting data stream is then transmitted over a secure and persistent WebSocket (Socket.IO) connection. On the server, data integrity is guaranteed through an atomic write mechanism, where the latest frame is written to a temporary file before being renamed, thus preventing downstream consumers in RealMirror from accessing incomplete data.

\subsection{OpenXR/Unity-Based Hand Gesture Acquisition}
\begin{figure}[H]
\centering
\includegraphics[width=0.8\linewidth]{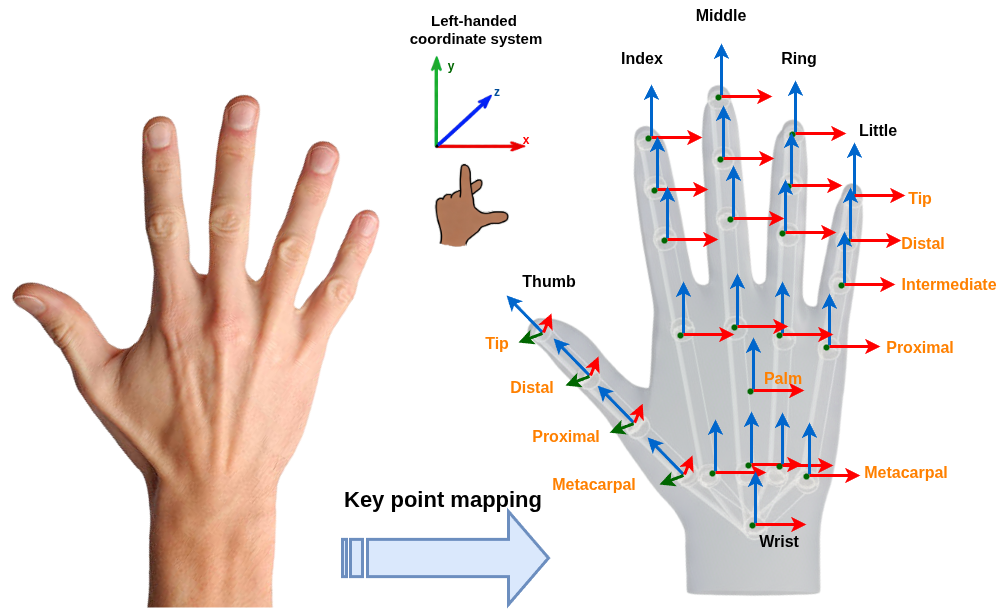}
\caption{The mapping and coordinate system representation of hand tracking.}
\label{fig:hand_track}
\end{figure}

To capture fine-grained hand motions, we leverage PICO’s native OpenXR APIs within the Unity engine. This path provides per-frame 3D position and orientation for all 26 tracked joints of each hand, enabling the system to represent dexterous manipulation. The mapping of these joints and their native left-handed coordinate system is illustrated in Figure \ref{fig:hand_track}. A key architectural choice is the decoupling of data sampling from the rendering loop. We enforce a fixed 60 Hz transmission rate, which ensures temporally consistent data intervals. This stability is crucial for both reliable real-time teleoperation and the construction of reproducible, high-quality datasets.

The coordinate transformation for the gesture stream is more complex than that of the handle stream. As noted, poses from the PICO Unity/OpenXR path are reported in a left-handed, Y-up coordinate system. Therefore, they first undergo a handedness normalization to a right-handed system before being remapped to IsaacSim’s Z-up convention. For network transport, each hand's joint data is serialized into a compact binary payload and transmitted via UDP. The choice of UDP as the transport protocol is motivated by the stream's high-frequency nature, which allows for tolerance to occasional packet loss. This trade-off is well-suited for real-time control applications.

\subsection{Reliability, Security, and Resource Efficiency}
\begin{itemize}
\item Reliability: The WebXR path re-establishes reference spaces on loss and handles session termination cleanly. The server persists only the latest frame and performs atomic writes to eliminate partial-file reads.

\item Security: WebXR runs in a secure context (HTTPS), and the transport uses authenticated TLS, satisfying browser policy and reducing attack surface in LAN deployment.

\item Efficiency: Minimal rendering eliminates GPU load during acquisition; quantization reduces bandwidth and numerical jitter; UDP binary streaming for hands minimizes serialization overhead.
\end{itemize}

\subsection{Data Post-processing}
In the field of robot teleoperation, end-effector jitter and sudden jumps are common issues. Specifically, one cause of jitter stems from biological tremor, meaning the human hand cannot remain completely stationary in space, resulting in inherent jitter in the raw data \citep{dai2020least}. Another cause of jitter is related to the nature of the Inverse Kinematics (IK) solver. When the positional displacement between two consecutive calculations is too small, the IK solution may still produce noticeable jitter in the end-effector \citep{buss2004introduction}.

Regarding sudden jumps, there are also two primary causes. Firstly, unstable raw data transmission or fluctuations in hand pose estimation can lead to significant jumps in the data. Secondly, the IK solver may encounter singularities or multiple solution scenarios, causing large, abrupt jumps in the end-effector's position \citep{guenther2008new}.

To effectively mitigate control instability induced by jitter and jumps, we designed a series of rules for kinematic optimization during the IK solving stage. This is achieved by filtering potential end-effector jitter and jumps between consecutive frames.

These related parameters are defined as: $t$ represents the current frame time index; $\mathbf{p}_t$ represents the position vector of the original end-effector of the current frame (from PICO parsing); $\boldsymbol{\theta}_t = [\theta_{t,1}, \theta_{t,2}, \dots, \theta_{t,m}]$ represents the original joint angle vector of the current frame (joint1 to jointm);  $\boldsymbol{\phi}_t = [\phi_{t,1}, \phi_{t,2}, \dots, \phi_{t,m}]$ represents the joint angle control signal vector after IK calculation in the current frame; The end position change vector is represented by the Euclidean distance between adjacent frames of the end joint: $\Delta \mathbf{p}_t = \mathbf{p}_t - \mathbf{p}_{t-1}$, $d_t = \| \Delta \mathbf{p}_t \|$; $\Delta \theta_{t,i} = | \theta_{t,i} - \theta_{t-1,i} |$ represents the absolute value of the inter frame change in the $i$ original joint angle $(i=1,..., m)$; $\Delta \phi_{t,i} = | \phi_{t,i} - \phi_{t-1,i} |$ represents the absolute value of the inter frame variation of the $i$ IK solved joint angle $(i=1,..., m)$; $\delta_1$ represents the first layer jitter filtering threshold (end distance); $\delta_2$ represents the second layer jitter filtering threshold (end distance, used for IK solvers); $\epsilon_1$ represents the third-layer jump filtering threshold (original joint angle); $\epsilon_2$ represents the fourth layer jump filtering threshold (IK calculated joint angle).

The filtering rules are defined as follows:
\begin{align}
\quad d_t \geq \delta_1 & \text{ (First-layer Jitter Filter)} \\
\quad d_t \geq \delta_2 & \text{ (Second-layer Jitter Filter)}\\
\quad \max_{i=1}^m \Delta \theta_{t,i} \leq \epsilon_1 & \text{ (Third-layer Jump Filter (Raw Data))}\\
\quad \max_{i=1}^m \Delta \phi_{t,i} \leq \epsilon_2 & \text{ (Fourth-layer Jump Filter (IK Solution))}
\end{align}

Executable data flag $E_t$ indicates whether the current frame data is executable:
\begin{equation} \label{eq:executable_condition}
E_t = \begin{cases} 
1 & \text{when } \left( d_t \geq \delta_1 \right) \land \left( d_t \geq \delta_2 \right) \land \left( \max_{i=1}^m \Delta \theta_{t,i} \leq \epsilon_1 \right) \land \left( \max_{i=1}^m \Delta \phi_{t,i} \leq \epsilon_2 \right), \\
0 & \text{otherwise,}
\end{cases}
\end{equation}
Through the aforementioned kinematic optimization, jitter and sudden jumps of the end-effector during teleoperation can be effectively suppressed. This optimization is natively integrated into the RealMirror platform, enabling optimal control performance whether using a handle or gesture-based input.

\section{Integration with RealMirror VLA Ecosystem
}

\begin{figure}[H]
\centering
\includegraphics[width=\textwidth]{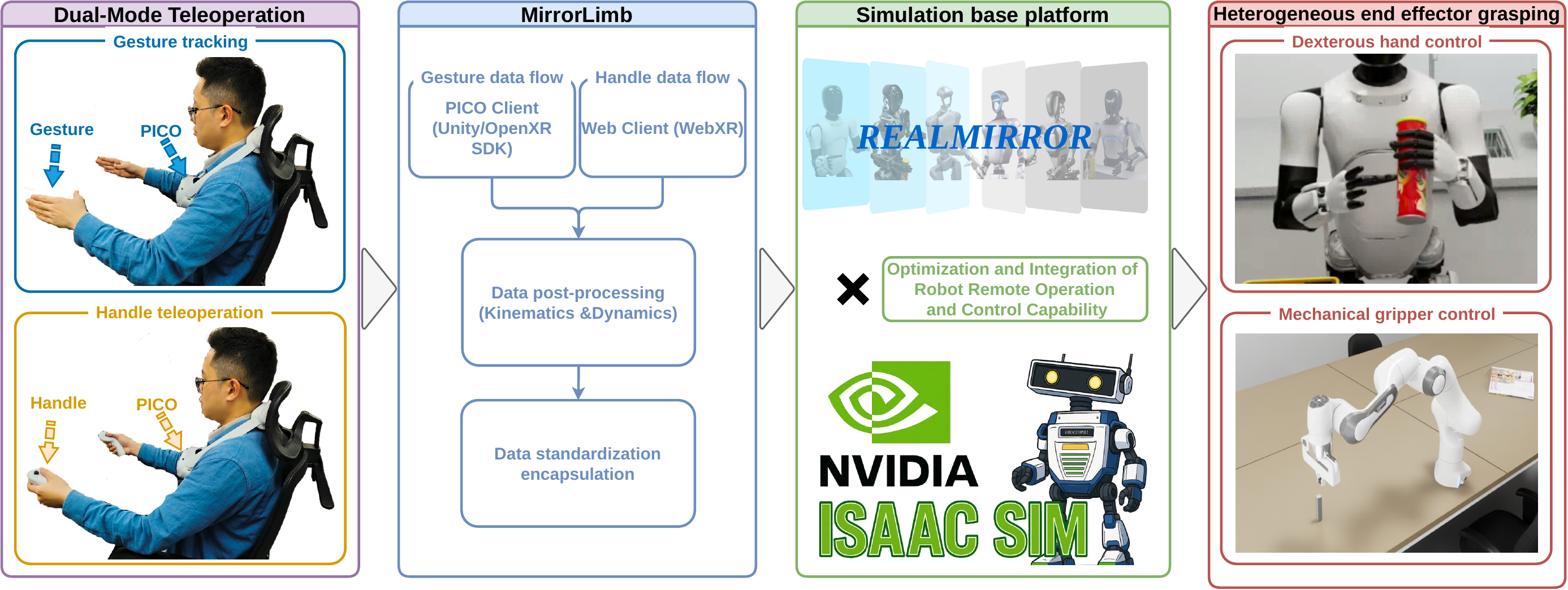}
\caption{Robot teleoperation system framework based on RealMirror ecosystem.}
\label{fig:first}
\end{figure}

To ensure broad applicability and optimal performance of MirrorLimb, we have natively integrated its gesture and handle-based teleoperation capabilities into RealMirror. Furthermore, we have standardized and exposed the MirrorLimb data interfaces. The complete end-to-end teleoperation workflow is illustrated in Figure \ref{fig:first}.
The workflow incorporates two selectable teleoperation modes. Empirical findings indicate that handle-based operation yields optimal data acquisition efficiency for simple tasks involving repetitive motions, whereas gesture proves more advantageous for complex tasks demanding continuous hand pose adjustments during data collection. The MirrorLimb system processes acquired gesture or handle data streams through subsequent data post-processing and kinematic optimization before integration into the RealMirror platform. On the robot side, kinematic constraints and dynamic optimizations have also been implemented to enhance compatibility with the MirrorLimb data stream, thereby achieving more stable robot control.
\subsection{Structured Gesture Data Flow}
To interface the teleoperation data flow with RealMirror, we have designed a standardized data interface. The structure of the gesture data stream is summarized in Table \ref{tab:hand_data_structure}.

It is worth noting that each recognizable joint is represented by position coordinates and a quaternion to convey its pose and orientation. The "handedness" label indicates whether the data corresponds to the left or right hand. The pose of the "Wrist" joint is selected as the representative pose for the entire hand.
\begin{table}[h!]
    \centering 
    \begin{tabular}{lll}
        \toprule
        \textbf{Field Name} & \textbf{Data Type} & \textbf{Description} \\
        \midrule

        \textbf{(Root)} & \textbf{Object} & \textbf{The root object representing a single data frame.} \\
        \hspace{1em}\texttt{timestamp} & Integer & Unix timestamp of the data frame. \\
        \hspace{1em}\texttt{data} & Array$<$Object$>$ & An array containing one or more `Hand` objects. \\
        \midrule

        \hspace{1em}\textbf{Hand Object} & \textbf{Object} & \textbf{Describes the complete information of a single hand.} \\
        \hspace{2em}\texttt{id} & String & Unique identifier for the hand in the current frame. \\
        \hspace{2em}\texttt{handedness} & String & Handedness ('left' or 'right'). \\
        \hspace{2em}\texttt{pose} & Object & The overall pose object of the hand. \\
        \hspace{3em}\texttt{position} & Array$<$Float$>$ & 3D position `[x, y, z]`. \\
        \hspace{3em}\texttt{orientation} & Array$<$Float$>$ & Orientation as a quaternion `[x, y, z, w]`. \\
        \hspace{2em}\texttt{joints} & Array$<$Object$>$ & An array of 26 `Joint` objects. \\
        \midrule

        \hspace{2em}\textbf{Joint Object} & \textbf{Object} & \textbf{Describes a single hand joint.} \\
        \hspace{3em}\texttt{name} & String & Name of the joint (e.g., 'Palm', 'ThumbTip'). \\
        \hspace{3em}\texttt{position} & Array$<$Float$>$ & The joint's 3D position `[x, y, z]`. \\
        \hspace{3em}\texttt{orientation} & Array$<$Float$>$ & The joint's orientation as a quaternion `[x, y, z, w]`. \\

        \bottomrule
    \end{tabular}
    \caption{Gesture tracking structured data.}
    \label{tab:hand_data_structure}
\end{table}

\subsection{Structured Handle Data Flow}
In contrast to the gesture data flow, while maintaining the same basic structure and key names, the handle data flow contains several unique parameters. The "profiles" field specifies a list of supported handle profiles or hardware identifiers, such as "oculus-touch-v2", "pico-phoenix", "pico-neo3", "pico-neo2", "oculus-touch", and "generic-trigger-squeeze-thumbstick". Furthermore, the extensive button information on the handle must be transmitted via the data stream to ensure precise handle-based teleoperation. This is particularly crucial for fine-tuning the robot wrist's roll, yaw, and pitch during teleoperation, which is achieved through specific handle buttons. The complete structure of the handle data flow is detailed in Table \ref{tab:controller_data_structure}.
\begin{table}[h!]
    \centering 
    \begin{tabularx}{\textwidth}{l l >{\raggedright\arraybackslash}X}
        \toprule
        \textbf{Field Name} & \textbf{Data Type} & \textbf{Description} \\
        \midrule

        \textbf{(Root)} & \textbf{Object} & \textbf{The root object for a single input data frame.} \\
        \hspace{1em}\texttt{timestamp} & Integer & Unix timestamp of the data frame. \\
        \hspace{1em}\texttt{data} & Array$<$Object$>$ & An array containing one or more \texttt{handle} objects. \\
        \midrule

        \hspace{1em}\textbf{handle Object} & \textbf{Object} & \textbf{Describes a single input handle.} \\
        \hspace{2em}\texttt{id} & String & Unique identifier for the handle (e.g., `right`). \\
        \hspace{2em}\texttt{handedness} & String & Specifies the handedness (`left` or `right`). \\
        \hspace{2em}\texttt{profiles} & Array$<$String$>$ & A list of supported handle profiles or hardware identifiers. \\
        \hspace{2em}\texttt{buttons} & Array$<$Object$>$ & An array of \texttt{Button State} objects, indexed by button type. \\
        \hspace{2em}\texttt{axes} & Array$<$Float$>$ & An array representing the state of analog axes (e.g., thumbsticks). \\
        \hspace{2em}\texttt{pose} & Object & The physical pose of the handle in 3D space. \\
        \hspace{3em}\texttt{position} & Array$<$Float$>$ & 3D position as `[x, y, z]`. \\
        \hspace{3em}\texttt{orientation} & Array$<$Float$>$ & Orientation as a quaternion `[x, y, z, w]`. \\
        \hspace{2em}\texttt{targetRayPose} & Object & The pose of the handle's targeting ray for pointing. \\
        \hspace{3em}\texttt{position} & Array$<$Float$>$ & 3D position of the ray's origin. \\
        \hspace{3em}\texttt{orientation} & Array$<$Float$>$ & Direction of the ray represented as a quaternion. \\
        \midrule

        \hspace{2em}\textbf{Button State Object} & \textbf{Object} & \textbf{Describes the state of a single button.} \\
        \hspace{3em}\texttt{pressed} & Boolean & True if the button is currently being fully pressed. \\
        \hspace{3em}\texttt{touched} & Boolean & True if the button is being touched (for capacitive sensors). \\
        \hspace{3em}\texttt{value} & Float & The analog value of the button/trigger, typically from 0.0 to 1.0. \\

        \bottomrule
    \end{tabularx}
    \caption{Handle tracking structured data.}
    \label{tab:controller_data_structure}
\end{table}

\section{Conclusion}
In this work, we have presented MirrorLimb, a cost-effective and robust framework for real-time hand pose acquisition and robot teleoperation, built upon the RealMirror ecosystem. By leveraging PICO XR devices, MirrorLimb effectively addresses the limitations of existing motion capture and visual tracking solutions, offering high-precision data acquisition through dual-channel streams based on WebXR and OpenXR protocols. The system natively integrates with IsaacSim, enabling stable trajectory recording and dexterous manipulation for VLA dataset construction. Through kinematic optimization, we effectively mitigate end-effector jitter and sudden jumps, ensuring reliable teleoperation performance. The standardized data interfaces for both gesture and handle inputs facilitate seamless adaptation to diverse robotic end-effectors, lowering the barrier for upper-limb manipulation research. 

MirrorLimb's compatibility with RealMirror not only accelerates VLA-related advancements but also provides a scalable platform for future explorations in embodied AI. Moving forward, we aim to extend support to additional XR devices and enhance real-time adaptability for complex manipulation tasks.
\bibliography{colm2024_conference}
\bibliographystyle{colm2024_conference}

\end{document}